\documentclass{article}

\PassOptionsToPackage{numbers}{natbib}

\usepackage{amsmath}
\usepackage{enumitem}
\usepackage{placeins}
\usepackage{needspace}
\usepackage{graphicx} 
\usepackage[
  colorlinks=true,
]{hyperref}
\usepackage{cleveref}
\usepackage{wrapfig}
\usepackage{makecell}
\usepackage{caption}
\usepackage[acronym]{glossaries}
\newacronym{sota}{SOTA}{state-of-the-art}
\newacronym{mlp}{MLP}{Multilayer Perceptron}
\newacronym{vit}{ViT}{Vision Transformer}
\newacronym{gcvit}{GC ViT}{Global Context Vision Transformer}
\newacronym{cnn}{CNN}{Convolutional Neural Network}
\newacronym{swin}{SwinT}{Shifted Window Transformer}
\newacronym{mmgcvit}{MICViT}{Multimodal Intra- and Cross-Context Vision Transformer}

\newacronym{mae}{MAE}{Mean Absolute Error}
\newacronym{mse}{MSE}{Mean Squared Error}
\newacronym{mri}{MRI}{Magnetic Resonance Imaging}

\newacronym{sl}{SL}{Separated Local}
\newacronym{sg}{SG}{Separated Global}
\newacronym{cl}{CL}{Cross Local}
\newacronym{cg}{CG}{Cross Global}

\newacronym{flops}{FLOPs}{Floating Point Operations}

\newacronym{adni}{ADNI}{Alzheimer's Disease Neuroimaging Initiative}
\newacronym{nako}{NAKO}{German National Cohort}
\newacronym{ukbb}{UKBB}{UK Biobank}
\newacronym{nlp}{NLP}{natural language processing}

\newacronym{bids}{BIDS}{Brain Imaging Data Structure}
\newacronym{fmri}{fMRI}{Functional MRI}
\newacronym{dti}{DTI}{Diffusion Tensor Imaging}
\newacronym{t1}{T1}{T1-weighted MRI}
\newacronym{t2}{T2}{T2-weighted MRI}
\newacronym{dwi}{DWI}{Diffusion-weighted imaging}
\newacronym{flair}{FLAIR}{Fluid-Attenuated Inversion Recovery}
\newacronym{swi}{SWI}{Susceptibility Weighted Imaging}

\newacronym{mhsa}{MHSA}{Multi-Head Self-Attention}
\newacronym{cgquery}{CG query}{cross-modal global query generator}
\newacronym{sgquery}{SG query}{separated global query generator}

\newacronym{rpe}{RPE}{Relative Position Encoding}

\newacronym{soop}{SOOP}{Stroke Outcome Optimization Project}
\newacronym{camcan}{Cam-CAN}{Cambridge Centre for Ageing and Neuroscience}

\newacronym{lf}{LF}{Late Fusion}
\newacronym{ef}{EF}{Early Fusion}
\glsdisablehyper

\newlength{\paperwrapwidth}
\newcommand{\eg}{\textit{e.g., }}

\usepackage[preprint]{neurips_2026}

\usepackage[utf8]{inputenc} 
\usepackage[T1]{fontenc}    
\usepackage{url}            
\usepackage{booktabs}       
\usepackage{amsfonts}       
\usepackage{nicefrac}       
\usepackage{microtype}      
\usepackage{xcolor}         

\title{Modeling Local, Global, and Cross-Modal Context in Multimodal 3D MRI}
\author{
  \textbf{Minh Duc Do\textsuperscript{1$^\ast$$^\dagger$}},
  \textbf{Tillmann Rheude\textsuperscript{1$^\ast$$^\dagger$}}, \\
  \textbf{Noel Kronenberg\textsuperscript{1}},
  \textbf{Roland Eils\textsuperscript{1,2,3,4$^\dagger$}},
  \textbf{Benjamin Wild\textsuperscript{1$^\dagger$}} \\
  \textsuperscript{1}Berlin Institute of Health at Charité - Universitätsmedizin Berlin\\
  \textsuperscript{2}Health Data Science Unit, Heidelberg University Hospital and BioQuant\\
  \textsuperscript{3}Intelligent Medicine Institute, Fudan University\\
  \textsuperscript{4}Department of Mathematics and Computer Science, Freie Universität Berlin \\
  \textsuperscript{$^\ast$ Equal contribution \quad $^\dagger$ Corresponding author}
}
\begin{document}
\maketitle

\begin{abstract}
Brain MRI poses a fundamental challenge for machine learning: models must learn from high-dimensional 3D data spanning multiple co-registered modalities, despite the limited sample sizes typical of neuroimaging studies relative to the diversity in anatomy, pathology, and acquisition conditions. While multimodal imaging provides complementary information critical for clinical interpretation, effectively integrating these signals remains difficult.
We propose \textbf{M}ultimodal \textbf{I}ntra- and \textbf{C}ross-Context \textbf{Vi}sion \textbf{T}ransformer (MICViT), a 3D vision transformer that explicitly models both modality-specific representations and cross-modal interactions across local and global contexts. Concretely, MICViT combines four attention mechanisms: modality-specific local and global attention for intra-modal feature learning, and cross-modal local and global attention to capture interactions between modalities.
We evaluate MICViT on brain age prediction across three heterogeneous datasets (UK Biobank, $n=41,404$; SOOP, $n=1,062$; Cam-CAN, $n=613$) using multiple MRI modalities (\eg T1, FLAIR, DWI, SWI). MICViT consistently outperforms state-of-the-art CNN and transformer baselines in 3D settings. Notably, it benefits more strongly from multimodal inputs, yielding larger performance gains as additional modalities are incorporated.
These results demonstrate that explicitly modeling intra- and cross-modal interactions is key to unlocking the full potential of multimodal brain MRI, highlighting a promising direction for representation learning in neuroimaging.

\end{abstract}

\section{Introduction}
Multimodal neuroimaging provides radiologists with non-invasive, high-resolution insights into the human brain, facilitating the identification of pathologies and supporting precise clinical interventions. 
In radiological practice, image interpretation is inherently multi-scale and multimodal: localized findings within a specific brain region are not assessed in isolation but are continuously contextualized with the global brain structure within the same modality, as well as with corresponding regions across other MRI modalities \cite{gilliesRadiomicsImagesAre2016}.
In diagnostic tasks, a single MRI modality (e.g., \gls{t1}) can provide valuable information about anatomical structures. However, its diagnostic potential is substantially enhanced when combined with additional modalities. For instance, \gls{t2} is sensitive to increased water content, making it particularly effective for visualizing edema and inflammation, \gls{fmri} captures brain activity, \gls{dwi} characterizes white matter tracts, and \gls{swi} reveals microbleeds and iron deposits \cite{haackeMagneticResonanceImaging1999}. In clinical practice, radiologists routinely integrate such complementary information to improve diagnostic accuracy.
Similarly, machine learning models that effectively combine multimodal brain MRI data can achieve more robust and accurate predictions \cite{schulzPerformanceReservesBrainimagingbased2024}. 

\begin{wrapfigure}[28]{rt}{\paperwrapwidth}
    \centering
    \vspace{2.6pt}
    \includegraphics[width=\linewidth]{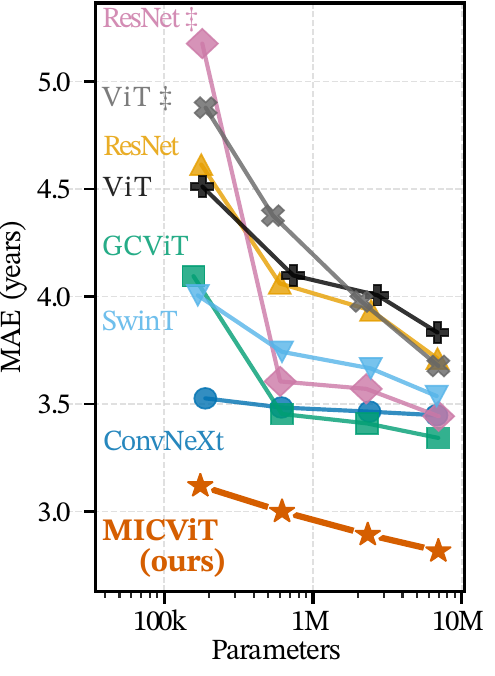}
    \caption{
        Scaling behavior on UKBB as a function of model size using MRI modalities (T1, FLAIR, SWI, DWI).
        MICViT consistently achieves the lowest error across datasets. Late fusion variants marked with $\ddag$
    }
    \label{fig:scaling_config}
\end{wrapfigure} 
Recent advances in transformer-based architectures have demonstrated strong potential for modeling long-range dependencies in medical imaging \cite{chenTransUNetTransformersMake2021}. In particular, \glspl{vit} enable flexible context aggregation beyond the locality constraints of \glspl{cnn} \cite{takahashiComparisonVisionTransformers2024}.
However, existing approaches face two key limitations when applied to multimodal 3D MRI data.

First, while prior work has explored cross-modal interactions, many architectures do not explicitly disentangle intra-modality and cross-modality processing within a unified and structured 3D MRI architecture. Early fusion strategies often entangle modality-specific signals, while late fusion approaches fail to fully exploit cross-modal dependencies \cite{huangFusionMedicalImaging2020}. This is limiting given that MRI modalities exhibit inherent spatial correspondence, providing complementary contrasts that enable precise voxel-wise comparison. At the same time, radiologists interpret local findings in the context of global brain anatomy, an interplay that is not explicitly modeled in prior work. Second, while unimodal 2D architectures have demonstrated the importance of jointly modeling local and global context, such designs are rarely extended to multimodal 3D MRI, limiting the ability to capture fine-grained anatomical structures alongside global patterns across modalities \cite{liuSwinTransformerHierarchical2021, hatamizadehGlobalContextVision2023}. This principle extends beyond healthcare; for example, in autonomous driving, the fusion of camera, LiDAR, and radar data enables more reliable perception under challenging environmental conditions \cite{chenMultiview3DObject2017}. 

Motivated by clinical radiological reasoning, our approach explicitly models intra- and cross-modality interactions across both local and global contexts within a unified architecture. We introduce \gls{mmgcvit}, a novel multimodal vision transformer for 3D brain MRI that captures how radiologists integrate localized findings with global brain context across modalities. While inspired by the unimodal \gls{gcvit} \cite{hatamizadehGlobalContextVision2023} and its combination of 2D image-based local self-attention with global context aggregation, our model incorporates four dedicated attention mechanisms for multimodal 3D MRI data: \gls{sl} and \gls{sg} for modality-specific representation learning, as well as \gls{cl} and \gls{cg} for explicit cross-modal interaction, enabling joint modeling of local structure, global context, and cross-modal dependencies.
Experiments across multiple neuroimaging datasets, including the large-scale \gls{ukbb}, suggest that performance in multimodal 3D neuroimaging is not solely determined by model size or data scale, but critically depends on how spatial and cross-modal interactions are modeled.

The main contributions of this work are as follows:
\begin{itemize} 
    \item We introduce \gls{mmgcvit}, a novel architecture for multimodal 3D brain MRI that jointly models intra- and cross-modal interactions through \gls{sl}, \gls{sg}, \gls{cl}, and \gls{cg}.
    \item We show that \gls{mmgcvit} achieves strong performance in a 3D setting, consistently outperforming established 3D baselines.
    \item We demonstrate that adding modalities yields larger relative performance gains for \gls{mmgcvit} than for baseline architectures.
    \item We provide a comprehensive empirical evaluation across multiple neuroimaging datasets and model scales, including large-scale ablations and systematic comparisons to parameter-matched baseline architectures.
\end{itemize}
\newpage

\section{Related Work}

\paragraph{Multimodal Learning in Brain MRI}
Multimodal learning has demonstrated improved performance compared to unimodal learning across domains such as computer vision, \gls{nlp}, and healthcare \cite{ngiamMultimodalDeepLearning2011,baltrusaitisMultimodalMachineLearning2017,sunMedicalMultimodalFoundation2024}.
Combining information from different modalities may contribute to enriched representations. This is especially pronounced in multimodal medical imaging, where different imaging modalities capture distinct yet complementary aspects of biological systems \cite{huangFusionMedicalImaging2020,luoMultimodalFusionBrain2024,liReviewDeepLearningbased2024,liemPredictingBrainageMultimodal2017,rokickiMultimodalImagingImproves2021,jirsaraieSystematicReviewMultimodal2023,mazherGeneralisableFoundationModels2025,liuMultimodalMRIVolumetric2022}.
For example, multimodality has been studied across structural MRI, functional MRI, diffusion imaging, and PET, with the goal of extracting both intra-modality and inter-modality information \cite{luoMultimodalFusionBrain2024}.
Notably, prior work has shown that leveraging multiple imaging modalities can yield performance gains comparable to substantially increasing dataset size \cite{schulzPerformanceReservesBrainimagingbased2024}.
More recently, foundation models based on transformer architectures trained on large-scale datasets \cite{takGeneralizableFoundationModel2026,mazherGeneralisableFoundationModels2025,sunMedicalMultimodalFoundation2024} aim to unify multimodal inputs within a single architecture, leveraging self-attention to capture complex cross-modal relationships.
Nevertheless, many existing methods rely on standard fusion schemes or shared encoders, and do not fully exploit the hierarchical, spatially co-registered, and volumetric dependencies present in multimodal brain MRIs.

\paragraph{Convolutional Neural Networks for Brain MRI Analysis}
\glspl{cnn} have dominated medical image analysis due to their strong inductive biases, such as locality and translation invariance, enabling robust performance on structured data like volumetric MRI \cite{kayhanTranslationInvarianceCNNs2020, mienyeDeepConvolutionalNeural2025}. 
Early work predominantly relied on 2D CNNs operating on individual MRI slices or patches due to the high computational cost of 3D MRI data \cite{guptaImprovedBrainAge2021, bashyamMRISignaturesBrain2020}. While this enables the use of 2D pretrained models, it ignores inter-slice dependencies, so anatomical relationships distributed across the 3D brain structure cannot be captured \cite{jonemoEfficientBrainAge2023}, limiting performance where spatial coherence is critical.
To address these limitations, subsequent work shifted toward 3D CNNs. 
Empirically, 3D approaches have achieved strong performance in neuroimaging tasks, including brain age prediction on large-scale datasets such as the UKBB \cite{pengAccurateBrainAge2021,dinsdaleLearningPatternsAgeing2021,leeDeepLearningbasedBrain2022,leonardsenDeepNeuralNetworks2022}.
However, most of these approaches remain unimodal, typically relying on a single imaging contrast such as T1-weighted MRI \cite{dinsdaleLearningPatternsAgeing2021}, thereby limiting their ability to exploit complementary information available in clinical multi-contrast protocols. 
More recent work extends CNNs to multimodal, multi-channel, or multi-stream MRI-derived representations to integrate complementary information. 
For example, attention-based fusion has been used to combine multiple MRI-derived channels for brain age estimation \cite{heMultichannelAttentionfusionNeural2021}, and \citet{liuMultimodalMRIVolumetric2022} introduce an attention-based fusion module for multimodal volumetric MRI.
Nevertheless, in many CNN-based models, multimodal interaction is confined to input concatenation or a dedicated fusion stage rather than being modeled explicitly and repeatedly throughout the network.

\paragraph{Vision Transformers in Medical Imaging} 
Transformers \cite{vaswaniAttentionAllYou2023,dosovitskiyImageWorth16x162021} have emerged as a powerful paradigm for many domains due to their ability to model long-range dependencies via self-attention. 
In medical imaging, there is a variety of attention-based architectures for, e.g., segmentation, classification, registration, and reconstruction \cite{chenTransUNetTransformersMake2021,liTransformingMedicalImaging2022}.
Early MRI attention-based methods often rely on 2D representations of volumetric data. For example, the Global-Local Transformer \cite{heGlobalLocalTransformerBrain2022} processes 2D slices using separate global and local pathways, but is limited to unimodal inputs and confines local-global interactions to explicit fusion modules. Similarly, the Triamese-ViT \cite{zhangTriameseViT3DAwareMethod2024} combines multiple 2D views to approximate 3D structure, reconstructing volumetric context only implicitly.
Recent work has shifted towards multimodal imaging. For example, \citet{zhaoTransformerBasedMultimodal2024} propose a multimodal attention-based fusion model.
However, cross-modal interactions are performed only after modality-specific encoding, limiting the modeling of fine-grained and hierarchical dependencies across modalities. Beyond structural MRI, SwiFT extends the \gls{swin} \cite{liuSwinTransformerHierarchical2021} to 4D spatiotemporal fMRI data using shifted-window attention \cite{kimSwiFTSwin4D2023}. However, it focuses on temporal dynamics rather than multimodal integration and relies on local attention with limited explicit modeling of global interactions. 
As another example, BrainFound processes 3D scans as sequences of 2D slices and supports multimodal inputs via channel-wise stacking \cite{mazherGeneralisableFoundationModels2025}. This slice-wise formulation again only approximates volumetric context and limits explicit modeling of fine-grained cross-modal interactions. Building on this, BrainIAC introduces a foundation model trained via contrastive self-supervised learning on large-scale multiparametric 3D MRI data \cite{takGeneralizableFoundationModel2026}, but modality-specific relationships remain implicitly modeled due to its shared encoder.
Building on this, existing approaches lack a unified architecture for multimodal imaging that jointly models full 3D spatial structure and explicitly captures both intra- and cross-modality interactions.

\section{Methods}
\begin{figure}[t]
    \centering
    \includegraphics[width=1\linewidth]{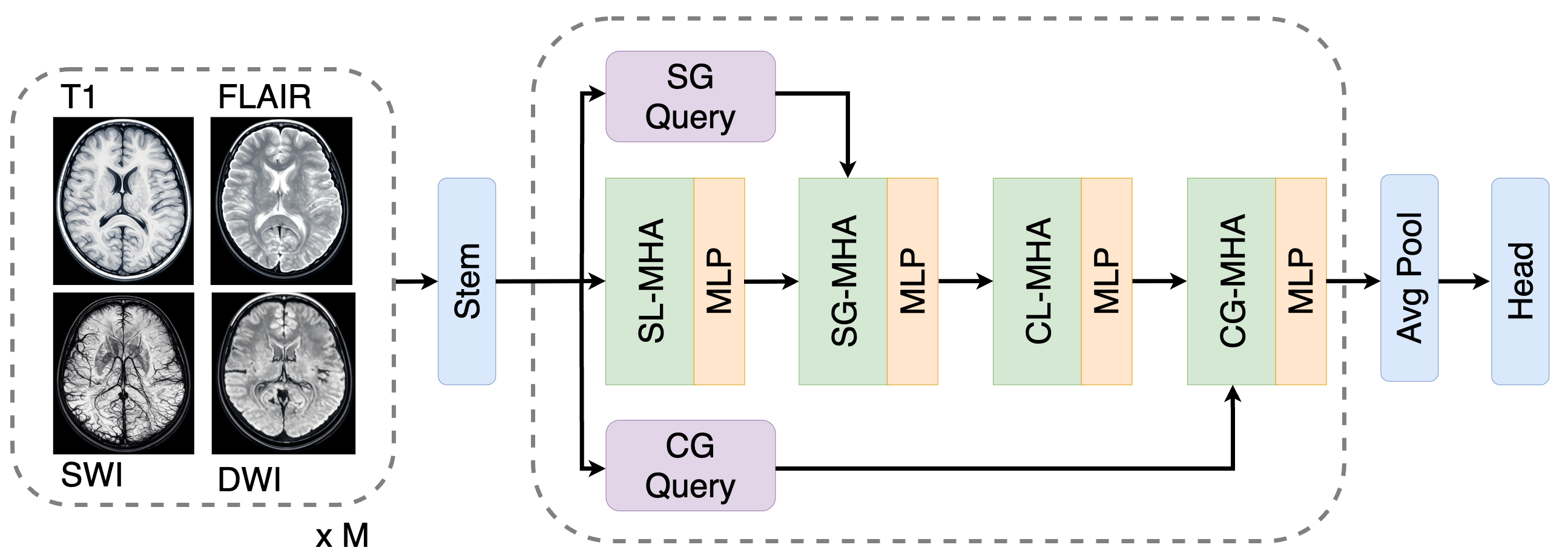}
        \caption{Architecture of the proposed \gls{mmgcvit}. Multiple imaging modalities (\eg T1, FLAIR, SWI, DWI) are processed via a stem that generates patch embeddings. The model sequentially applies Separated Local (SL) and Separated Global (SG) attention for intra-modality representation learning, followed by Cross Local (CL) and Cross Global (CG) attention to enable explicit cross-modality fusion. Dedicated global query generators are used for \gls{sg} and \gls{cg}.}
    \label{fig:architecture}
\end{figure}
\subsection{Implementation of MICViT}
Let $x \in \mathbb{R}^{B \times M \times H \times W \times D}$ denote a batch of volumetric inputs with $B$ samples, $M$ MRI modalities, and spatial resolution $H \times W \times D$. Our \gls{mmgcvit} extends the original \gls{gcvit} to jointly model modality-specific structure and cross-modal interactions in 3D neuroimaging (\Cref{fig:architecture}). Specifically, each modality is first processed independently by a modality-specific 3D patch embedding stem layer, producing an initial token tensor
$
z^{(0)} \in \mathbb{R}^{B \times M \times H_0 \times W_0 \times D_0 \times C_0},
$
where $C_0 = \mathrm{dim}$ and $(H_0, W_0, D_0)$ denote the patch-level spatial resolution (see \Cref{sec:stem} for details). The resulting embeddings are then partitioned into non-overlapping 3D windows of size $W_d \times W_h \times W_w$. The \gls{mmgcvit} alternates between four attention mechanisms designed to disentangle intra-modality and cross-modality dependencies at both local and global scales: \gls{sl}, \gls{sg}, \gls{cl}, and \gls{cg} attention. The model can, in principle, stack multiple blocks composed of these attention layers. However, preliminary experiments showed no meaningful performance improvement on the evaluated datasets when using deeper stacked architectures. To enable a controlled and systematic analysis, we therefore adopt a single-block architecture with four attention layers.
Our multimodal design employs two complementary query generators: (i) a \gls{sgquery} that extracts modality-specific global context independently per modality, and (ii) a \gls{cgquery} that fuses modalities along the channel dimension and projects the resulting shared representation back into the per-modality attention space. Formally, these modules produce $q_{\mathrm{sep}}$ and $q_{\mathrm{cross}}$, which parameterize the \gls{sg} and \gls{cg} blocks, respectively. This enables the model to retain modality-specialized representations while explicitly modeling cross-modal dependencies. Each transformer block follows a standard residual structure consisting of normalization, window-based attention, residual connection, and multilayer perceptron (MLP) update, with stochastic depth applied across layers. After the final block, the representation
$
z^{(1)} \in \mathbb{R}^{B \times M \times H_0 \times W_0 \times D_0 \times C_0}
$
is normalized and reshaped by merging the modality and channel dimensions to obtain
$
\tilde{z} \in \mathbb{R}^{B \times (M C_0) \times H_0 \times W_0 \times D_0}.
$
Global average pooling yields a compact subject-level embedding
$
h \in \mathbb{R}^{B \times M C_0},
$
which is fed into a multitask prediction head. Across model scales, we vary the base embedding width and number of attention heads while keeping the architectural structure fixed (\Cref{tab:model_configs}).

\subsection{Attention Types of MICViT}
\begin{figure}[t]
    \centering
    \includegraphics[width=1\linewidth]{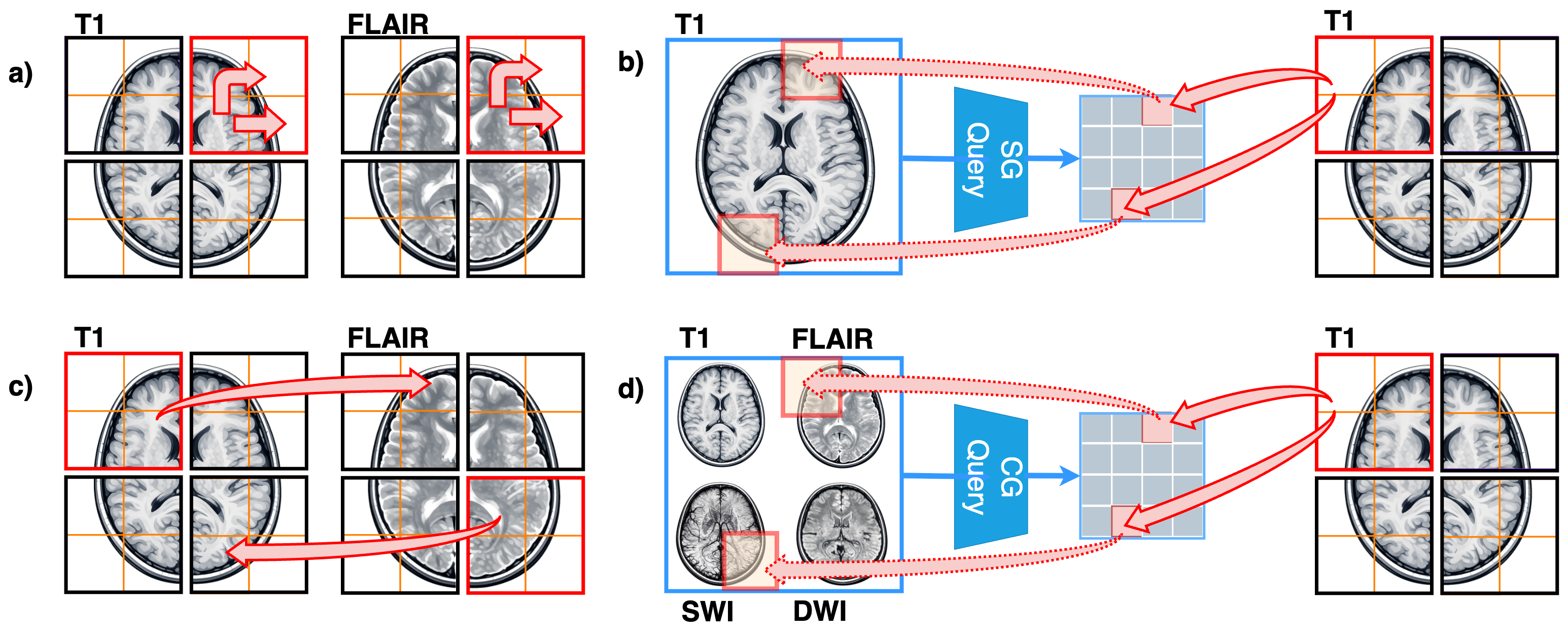}

       \caption{Illustration of the four attention types operating on latent 3D representations:
        (a) \gls{sl} performs local self-attention within spatial windows independently for each modality;
        (b) \gls{sg} aggregates global context within each modality via shared global queries;
        (c) \gls{cl} applies local cross-attention across modalities within corresponding spatial windows;
        (d) \gls{cg} models global cross-modal interactions using shared global queries across modalities.}
    \label{fig:attn_mechanics_2}
\end{figure}
\paragraph{Separated Local (SL) Attention.}
Local attention is applied independently within each modality, restricting interactions to spatially neighboring patches of size $W_d \times W_h \times W_w$ and preventing any cross-modal information exchange (\Cref{fig:attn_mechanics_2}a). For each modality $k \in \{1,\dots,M\}$ and window $w \in \mathcal{W}$, let $Z_w^{(k)} \in \mathbb{R}^{N_w \times C}$ denote the corresponding tokens. Queries, keys, and values are obtained via linear projections:
$
Q_w^{(k)}, K_w^{(k)}, V_w^{(k)} \in \mathbb{R}^{N_w \times C}.
$
Here, $C$ denotes the token dimension, and $d_h$ denotes the per-head channel dimension. Self-attention is then computed independently for each modality, window, and attention head $h$:
\begin{equation}
\text{Attention}_w(Q_w^{(k),h}, K_w^{(k),h}, V_w^{(k),h})
=
\text{softmax}\left(
\frac{Q_w^{(k),h} (K_w^{(k),h})^\top}{\sqrt{d_h}} + B_h^{\text{spatial}}
\right) V_w^{(k),h}.
\end{equation}

We incorporate a 3D relative positional encoding, where the bias $B_h^{\text{spatial}}$ is derived from a learnable table $\hat{B} \in \mathbb{R}^{(2W_d-1)\times(2W_h-1)\times(2W_w-1)}$, indexed by relative offsets along depth, height, and width. 

\paragraph{Separated Global (SG) Attention.}
To incorporate global contextual information, we extend the global attention mechanism of \gls{gcvit} to the multimodal 3D setting (\Cref{fig:attn_mechanics_2}b). While local window attention restricts interactions to spatial neighborhoods, \gls{sg} enables each window to access modality-specific global context via a shared set of global query tokens. For each modality $k \in \{1,\dots,M\}$, we generate a modality-specific global query tensor $Q_g^{(k)} \in \mathbb{R}^{N_q \times C}$ from the corresponding modality tokens. These queries are reshaped per head as $Q_g^{(k),h} \in \mathbb{R}^{N_q \times d_h}$ and shared across all local windows of the same modality. The global queries are broadcast to match the number of windows, yielding an effective batch size $B^* = B \times |\mathcal{W}|$. Importantly, the global queries are computed once per modality and sample, and then reused across all windows:
$
Q_{g,w}^{(k)} = Q_g^{(k)},\quad\forall w \in \mathcal{W}.
$
Keys and values are computed independently within each local window:
$
K_w^{(k),h}, V_w^{(k),h} \in \mathbb{R}^{N_w \times d_h}.
$
The global queries attend to local window tokens within the same modality:
\begin{equation}
\text{Attention}_w(Q_g^{(k),h}, K_w^{(k),h}, V_w^{(k),h})
=
\text{softmax}\left(
\frac{Q_g^{(k),h} (K_w^{(k),h})^\top}{\sqrt{d_h}} + B_h^{\text{spatial}}
\right) V_w^{(k),h}.
\end{equation}

\paragraph{Cross Local (CL) Attention.}
To explicitly model and fuse interactions between multiple 3D data modalities, we extend local window attention to jointly process tokens from all modalities within each spatial region (\Cref{fig:attn_mechanics_2}c). Given modality-specific tokens $Z^{(k)}$, $k \in \{1,\dots,M\}$, we construct a multimodal representation for each window $w$ by concatenating tokens across modalities:
$
Z_w = [Z_w^{(1)}, \dots, Z_w^{(M)}],
$
resulting in a sequence of $M N_w$ tokens. Attention is computed over this joint representation, enabling both intra-modality (self-attention within $Z_w^{(k)}$) and cross-modality (cross-attention across $Z_w^{(k)}$ and $Z_w^{(k')}$ for $k \neq k'$) interactions within each local window. Queries, keys, and values are obtained via linear projections:
$
Q_w, K_w, V_w \in \mathbb{R}^{(M N_w) \times C}.
$
To model both spatial and modality-dependent relationships, we employ a factorized relative position bias:
\begin{equation}
(B_h^{\text{multi}})_{i,j}
=
(B_h^{\text{spatial}})_{p_i,p_j}
+
(B_h^{\text{modality}})_{m_i,m_j},
\end{equation}
where $p_i$ and $m_i$ denote the spatial position within the window and the modality index of token $i$, respectively. The spatial component $B_h^{\text{spatial}} \in \mathbb{R}^{N_w \times N_w}$ is shared across modalities, while $B_h^{\text{modality}} \in \mathbb{R}^{M \times M}$ captures modality-specific interactions.

The attention operation for a window $w$ and head $h$ is given by:
\begin{equation}
\text{Attention}_w(Q_w^h, K_w^h, V_w^h)
=
\text{softmax}\left(
\frac{Q_w^h (K_w^h)^\top}{\sqrt{d_h}} + B_h^{\text{multi}}
\right) V_w^h.
\end{equation}

\paragraph{Cross Global (CG) Attention.}
Analogous to cross-local attention, we construct a multimodal window
$
Z_w = [Z_w^{(1)}, \dots, Z_w^{(M)}] \in \mathbb{R}^{(M N_w)\times C}.
$
In cross-global attention, queries are not modality-specific. Instead, modality features are concatenated along the channel dimension to generate a single shared global query tensor, which is projected per head and replicated across modalities, yielding identical modality slices. Thus, queries are shared, while keys and values are computed from the multimodal window tokens.

For each head \(h\), let
$
Q_g^h \in \mathbb{R}^{(M N_w)\times d_h}, \quad
K_w^h, V_w^h \in \mathbb{R}^{(M N_w)\times d_h}.
$
Attention is computed as
\[
\mathrm{Attention}_w(Q_g^h,K_w^h,V_w^h)
=
\mathrm{softmax}\!\left(
\frac{Q_g^h (K_w^h)^\top}{\sqrt{d_h}} + B_h^{\mathrm{multi}}
\right)V_w^h.
\]

\section{Experiments}

\paragraph{Datasets}
\begin{wraptable}[12]{r}{\paperwrapwidth}
\centering
\footnotesize
\setlength{\tabcolsep}{3pt}
\renewcommand{\arraystretch}{1.05}
\vspace{-5 pt}
\caption{Dataset characteristics across Cam-CAN, SOOP, and UK Biobank.}
\label{tab:dataset_stats}
\begin{tabular}{@{}lccc@{}}
\toprule
 & \textbf{Cam-CAN} & \textbf{SOOP} & \textbf{UKBB} \\
\midrule
$n$        & 613            & 1{,}062         & 41{,}404 \\
Women (\%) & 49.4           & 47.5            & 52.9 \\
Age (yr)   & 54.7$\pm$18.6  & 63.8$\pm$13.7   & 63.9$\pm$7.7 \\
\midrule
Modalities
& \makecell[c]{T1\\T2\\DWI\\MTI}
& \makecell[c]{T1\\FLAIR\\rec-TRACE\\rec-ADC}
& \makecell[c]{T1\\FLAIR\\SWI\\DWI} \\
\bottomrule
\end{tabular}
\end{wraptable}

We evaluate our approach on three complementary multimodal neuroimaging datasets (\Cref{tab:dataset_stats}): \gls{ukbb} \cite{millerMultimodalPopulationBrain2016, littlejohnsUKBiobankImaging2020}, \gls{soop} \cite{absherStrokeOutcomeOptimization2024}, and \gls{camcan} \cite{shaftoCambridgeCentreAgeing2014, taylorCambridgeCentreAgeing2017}. The \gls{ukbb} is a population-based cohort of 41,404 participants (age=63.9$\pm$7.7) with four imaging modalities, including \gls{t1}, \gls{flair}, \gls{swi}, and \gls{dwi}. In addition, we consider the \gls{soop}, a clinical dataset comprising 1,062 stroke patients (age=63.8$\pm$13.7) with multimodal brain MRI, including \gls{t1}, \gls{flair}, and diffusion-derived maps (a reconstructed trace-weighted diffusion MRI image and a reconstructed Apparent Diffusion Coefficient (ADC) map). Finally, we consider \gls{camcan}, which comprises 613 participants (age=54.7$\pm$18.6 years), with \gls{t1}, \gls{t2}, \gls{dwi}, and magnetization transfer imaging (MTI). \gls{camcan} is designed to investigate age-related changes in brain structure, function, and cognition, providing a controlled lifespan cohort that isolates biologically meaningful variation from confounding acquisition effects. MRI data were preprocessed using a standardized pipeline (Python 3.11, ANTs 2.5.1). MRI volumes were denoised and then underwent bias field correction, skull stripping, and intensity normalization. All scans were co-registered to MNI space to ensure spatial consistency. Due to the high computational cost of volumetric data, volumes were downsampled from $224^3$ to $64^3$ resolution unless stated otherwise (see \Cref{sec:mri_preprocessing} for details).

\paragraph{Optimization}
For training and evaluation, we follow best practices for multimodal learning \cite{rheudeFusionConfusionMultimodal2025}, including Schedule-Free Learning \cite{defazioRoadLessScheduled2024}. We further incorporate Gap-Guided Sharpness-Aware Minimization (GSAM) \cite{zhuangSurrogateGapMinimization2022a}, which we found to be essential for achieving stable convergence and improving generalization performance. Mimetic initialization is applied to all models \cite{trockmanMimeticInitializationSelfAttention2023} to strengthen inductive biases in self-attention. In addition, we employ pre-layer normalization within the transformer architecture to further enhance training stability \cite{xiongLayerNormalizationTransformer2020} and perform hyperparameter sweeps. 
Models are hyperparameter-optimized via grid search over learning rates from $10^{-2}$ to $10^{-5}$, with a fixed weight decay of $0.05$. Due to the high memory demands of 3D inputs, we use a batch size of $2$ and bfloat16 precision. Early stopping is applied based on validation loss. All experiments use 5-fold cross-validation on downsampled $64^3$ inputs.
To ensure fair and well-calibrated comparisons, all models are tuned via grid search. In addition, we perform extensive hyperparameter optimization for ConvNeXt \cite{liuConvNet2020s2022} using Bayesian sweeps with $100$ runs to establish a particularly strong reference baseline (\Cref{tab:hyperparams_convnext}). We conduct these sweeps across multiple model scales, including \emph{small}, \emph{medium}, \emph{base}, and \emph{large} configurations, to comprehensively explore the capacity-performance trade-off. The resulting best-performing configuration serves as a reference point for all subsequent comparisons.
Each dataset is independently partitioned into training, validation, and test splits using a 5-fold cross-validation scheme. In each fold, 80\% of the data is used for training, 10\% for validation, and 10\% for testing, ensuring subject-level separation across splits. Models are trained and evaluated separately on each dataset.
Model selection is performed based on the lowest validation loss, defined as the mean absolute error (MAE) on the validation set. All datasets are evaluated on the common task of brain age prediction, a standard benchmark for representation learning in neuroimaging. Performance is measured using mean absolute error (MAE) in years and reported as the mean and standard deviation over 5 folds.

\paragraph{Comparison Protocol}
We adopt a \emph{parameter-matched} protocol to compare models at equal representational capacity and additionally report wall-clock time per training epoch as a direct measure of realized runtime. We avoid FLOP-matching because FLOPs are an unreliable proxy for actual cost on modern hardware, where wall-clock time is shaped by memory access patterns and kernel-level efficiency rather than raw operation counts \cite{maShuffleNetV2Practical2018,daoFlashAttentionFastMemoryEfficient2022,inbarTimeMattersScaling2024}. This choice also sidesteps a structural issue specific to 3D: with $N \propto L^3$ tokens, ViT self-attention FLOPs scale as $\mathcal{O}(L^6 C)$ while parameters remain at $\mathcal{O}(C^2)$, whereas 3D CNN FLOPs scale as $\mathcal{O}(L^3 K^3 C_{\text{cnn}}^2 D_{\text{layer}})$ with parameters $\mathcal{O}(D_{\text{layer}} K^3 C_{\text{cnn}}^2)$, so matching FLOPs across families would require disproportionate CNN capacity inflation and make joint FLOP--parameter alignment infeasible.
\WFclear

\section{Results}

\Needspace{0.44\textheight}
\begin{wraptable}[28]{r}{\paperwrapwidth}
    \centering
    \setlength{\tabcolsep}{0.9pt}
    \renewcommand{\arraystretch}{0.96}

    \vspace{-5 pt}
    \caption{
        Ablation study of MRI modality combinations in the UKBB.
    }
    \footnotesize
    
    \begin{tabular}{@{}l r r r@{}}
    \toprule
    \textbf{Modalities}
    & \makecell{\textbf{MAE} $\downarrow$\\\textbf{$\pm$ SD}}
    & \makecell{
        \textbf{Params} \\
        \textbf{(M)} $\downarrow$
    }
    & \makecell{
	        \textbf{Time} \\
	        \textbf{(min/ep.)} $\downarrow$
    } 
    \\
    \midrule
    
    \multicolumn{4}{@{}l}{\textbf{GCViT}} \\
    \midrule
    SWI & $4.31 \pm 0.06$ & 2.28 & 30.42 \\
    T1  & $3.81 \pm 0.10$ & 2.28 & 28.02 \\
    FLAIR  & $3.44 \pm 0.05$ & 2.28 & 30.39 \\
    DWI & $3.62 \pm 0.07$ & 2.28 & 30.28 \\
    FLAIR+DWI & $3.34 \pm 0.05$ & 3.56 & 39.27 \\
    T1+FLAIR+DWI & $3.38 \pm 0.02$ & 5.01 & 61.46 \\
    all 4 modalities & $3.34 \pm 0.06$ & 6.96 & 86.33 \\
    
    \addlinespace[2pt]
    \midrule
    \multicolumn{4}{@{}l}{\textbf{ConvNeXt}} \\
    \midrule
    SWI & $4.51 \pm 0.03$ & 2.34 & 24.27 \\
    T1  & $3.97 \pm 0.06$ & 2.34 & 24.21 \\
    FLAIR  & $3.72 \pm 0.03$ & 2.34 & 24.41 \\
    DWI & $3.92 \pm 0.04$ & 2.34 & 24.29 \\
    FLAIR+DWI & $3.56 \pm 0.02$ & 3.58 & 35.64 \\
    T1+FLAIR+DWI & $3.48 \pm 0.07$ & 5.00 & 50.04 \\
    all 4 modalities & $3.53 \pm 0.03$ & 6.77 & 67.67 \\
    
    \addlinespace[2pt]
    \midrule
    \multicolumn{4}{@{}l}{\textbf{\gls{mmgcvit}}} \\
    \midrule
    SWI & $4.12 \pm 0.20$ & 2.28 & 30.73 \\
    T1  & $3.78 \pm 0.11$ & 2.28 & 30.93 \\
    FLAIR  & $3.28 \pm 0.10$ & 2.28 & 31.77 \\
    DWI & $3.17 \pm 0.02$ & 2.28 & 28.34 \\
    FLAIR+DWI & $3.08 \pm 0.09$ & 3.55 & 63.61 \\
    T1+FLAIR+DWI & $2.87 \pm 0.14$ & 5.04 & 75.74 \\
    all 4 modalities & $\mathbf{2.82} \pm 0.06$ & 6.97 & 89.45 \\
    \bottomrule
    \end{tabular}
    
    \label{tab:ablation_modalities}
\end{wraptable}

\paragraph{Performance in multimodal 3D MRI depends on modeling spatial and cross-modal interactions}
We first study how the combination and ordering of attention mechanisms affect performance in multimodal 3D MRI for brain age prediction. We train on \gls{t1}, \gls{flair}, \gls{swi}, and \gls{dwi}, where it is not obvious which information the model should prioritize at different stages of processing. We find that the sequence of intra- and cross-modal attention is critical. Evaluating all permutations of \gls{sl}, \gls{sg}, \gls{cl}, and \gls{cg} attention on \gls{ukbb}, we observe that the full configuration (\gls{sl}, \gls{sg}, \gls{cl}, \gls{cg}) consistently achieves the best performance (\Cref{tab:single_layer_attention_permutations}). This indicates that the gains do not arise from a single dominant component, but from their interaction.

\paragraph{MICViT scales effectively with an increasing number of MRI modalities}
Building on this, we next analyze how models benefit from increasing numbers of MRI modalities (\Cref{tab:ablation_modalities}). We evaluate single modalities, and select the best-performing pair (\gls{flair} + \gls{dwi}), as well as extensions to three and four modalities. In the unimodal setting, \gls{mmgcvit} also achieves the lowest MAE for every individual contrast, outperforming all baselines (\gls{gcvit} and ConvNeXt) for \gls{swi}, \gls{t1}, \gls{flair}, and \gls{dwi} (\Cref{tab:ablation_modalities}). We find that performance improves consistently as more modalities are added, confirming the benefit of complementary imaging information. In this setting, we demonstrate that \gls{mmgcvit} benefits most from the incremental addition of modalities, outperforming ConvNeXt as a strong \gls{cnn} baseline and \gls{gcvit} as a transformer baseline across all configurations. The performance gap increases with the number of modalities, indicating that \gls{mmgcvit} more effectively exploits cross-modal complementarities (\Cref{tab:model_configs_modalities_final}).

\paragraph{Robust performance across multiple datasets}
To evaluate robustness across heterogeneous data, all models are trained and evaluated independently across datasets that differ in sample size and acquisition characteristics (\Cref{tab:cross_dataset_results_final}). \gls{mmgcvit} consistently achieves the lowest MAE on all datasets, outperforming CNN-based (ResNet, ConvNeXt) and transformer-based (\gls{vit}, \gls{swin}, \gls{gcvit}) baselines. Improvements over strong baselines such as \gls{gcvit} and ConvNeXt are consistent, with more pronounced gains observed in smaller datasets. Performance improvements remain stable across folds. The stronger gains in low-data settings indicate improved sample efficiency. We attribute this to the architecture: decoupled intra-modality encoding (\gls{sl}, \gls{sg}) reduces interference, while explicit cross-attention (\gls{cl}, \gls{cg}) enables structured inter-modality fusion, which is particularly beneficial in heterogeneous and data-limited settings.

\Needspace{0.30\textheight}
\begin{wrapfigure}[15]{r}{\paperwrapwidth}
    \centering
    \vspace{-1 pt}
    \includegraphics[width=\linewidth]{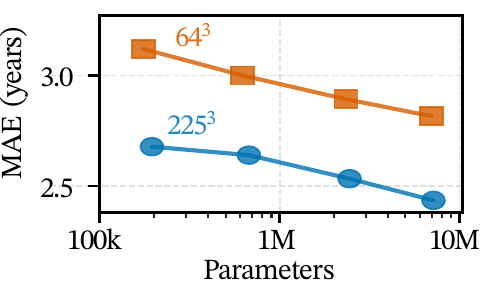}
    \caption{Scaling with model capacity and input resolution. Full-resolution MRI ($225^3$) reduces MAE across model sizes relative to downsampled inputs ($64^3$).}
    \label{fig:mmGCViT_mri_resolution}
\end{wrapfigure}
\paragraph{Consistent scaling gains from model capacity and input resolution}
Finally, we analyze scaling behavior across model capacity and input resolution. We show that \gls{mmgcvit} maintains superior performance across all scaling configurations, from small to large models (\Cref{fig:scaling_config}, \Cref{tab:model_configs}, \Cref{tab:scaling_results}). This advantage persists when increasing the parameter count, highlighting consistent and effective scaling behavior. We further demonstrate that higher input resolution leads to an additional performance gain (\Cref{fig:mmGCViT_mri_resolution}). While consistent improvements are already observed on downsampled inputs (MAE: $3.12 \rightarrow 2.82$ years), full-resolution inputs ($225^3$) shift performance to substantially lower error levels (MAE: $2.68 \rightarrow 2.44$ years), yielding improvements of up to $\sim 0.6$ years across model scales. Overall, these results show that both the structured design of attention mechanisms and their interaction with multimodal inputs and resolution are key drivers of performance in 3D brain MRI.

\paragraph{Combining global and local attention across and within modalities is critical for performance}
\Needspace{0.28\textheight}
\begin{wraptable}[17]{r}{\paperwrapwidth}
    \centering
    \setlength{\tabcolsep}{0.3pt}
    \renewcommand{\arraystretch}{1.0}

    \vspace{-5 pt}
       \caption{
        Dataset evaluation on SOOP, Cam-CAN, and UKBB (large model configuration, $64^3$ image resolution).
    }
    \footnotesize
    
    \begin{tabular}{l r r r}
    \toprule
    & \makecell{\textbf{SOOP} \\ (n=1,715)}
    & \makecell{\textbf{Cam-CAN} \\ (n=653)}
    & \makecell{\textbf{UKBB} \\ (n=41,404)} \\
    \cmidrule(lr){2-2} \cmidrule(lr){3-3} \cmidrule(lr){4-4}
    \textbf{Model} 
    & \makecell{\textbf{MAE} $\downarrow$\\\textbf{$\pm$ SD}}
    & \makecell{\textbf{MAE} $\downarrow$\\\textbf{$\pm$ SD}}
    & \makecell{\textbf{MAE} $\downarrow$\\\textbf{$\pm$ SD}} \\
    \midrule
    SwinT             & $9.13 \pm 0.23$ & $10.58 \pm 0.92$ & $3.54 \pm 0.08$ \\
    ResNet            & $8.83 \pm 1.49$ & $8.72 \pm 1.29$ & $3.71 \pm 0.17$ \\
    \gls{vit}         & $8.81 \pm 1.79$ & $8.94 \pm 1.05$ & $3.83 \pm 0.15$ \\
    ResNet $\ddag$    & $7.96 \pm 0.93$ & $8.57 \pm 1.17$ & $3.44 \pm 0.12$ \\
    ConvNeXt          & $7.33 \pm 0.55$ & $8.22 \pm 0.72$ & $3.45 \pm 0.04$ \\
    \gls{vit} $\ddag$ & $7.29 \pm 0.49$ & $9.12 \pm 0.82$ & $3.68 \pm 0.07$ \\
    \gls{gcvit}       & $\underline{6.76} \pm 0.14$ & $\underline{7.85} \pm 0.78$ & $\underline{3.34} \pm 0.07$ \\
    \midrule
    \textbf{\gls{mmgcvit}} & $\mathbf{6.67} \pm 0.10$ & $\mathbf{6.43} \pm 0.69$ & $\mathbf{2.82} \pm 0.06$ \\
    \bottomrule
    \end{tabular}
    \label{tab:cross_dataset_results_final}
\end{wraptable}

\begin{figure}[t]
    \centering
    \begin{minipage}[t]{0.51\linewidth}
        \vspace{0pt}
        \centering
\setlength{\tabcolsep}{1.3pt}
\renewcommand{\arraystretch}{1.02}

\captionof{table}{
    Ablation study on attention mechanisms.
    Individual and combined attention components are removed from \gls{mmgcvit}.
}
    
\begin{tabular}{@{}l r r r@{}}
    \toprule
    \textbf{Configuration}
    & \makecell{\textbf{MAE} $\downarrow$\\\textbf{$\pm$ SD}}
    & \makecell{
        \textbf{Params} \\ 
        \textbf{(M)} $\downarrow$
    }
    & \makecell{
	        \textbf{Time} \\
	        \textbf{(min/ep.)} $\downarrow$
    } 
    \\
    \midrule
    
    Full \gls{mmgcvit} & $3.12 \pm 0.19$ & 0.18 & 38.84 \\
    
    \addlinespace[2pt]
    \midrule
    w/o SL & $3.14 \pm 0.21$ & 0.16 & 43.30 \\
    w/o CL & $3.30 \pm 0.31$ & 0.16 & 43.31 \\
    w/o SG & $3.32 \pm 0.46$ & 0.16 & 43.30 \\
    w/o CG & $3.33 \pm 0.24$ & 0.16 & 43.60 \\
    
    \addlinespace[2pt]
    \midrule
    w/o SL--SG & $3.32 \pm 0.46$ & 0.15 & 35.97 \\
    w/o SL--CL & $3.44 \pm 0.26$ & 0.15 & 38.97 \\
    w/o CL--CG & $3.45 \pm 0.27$ & 0.15 & 39.27 \\
    w/o SG--CL & $3.54 \pm 0.45$ & 0.15 & 40.42 \\
    w/o SG--CG & $4.06 \pm 0.20$ & 0.15 & 29.03 \\
    
    \addlinespace[2pt]
    \midrule
    w/o SL--SG--CL & $3.41 \pm 0.49$ & 0.14 & 43.62 \\
    w/o SL--SG--CG & $4.28 \pm 0.39$ & 0.14 & 44.91 \\
    w/o SG--CL--CG & $4.35 \pm 0.14$ & 0.14 & 45.08 \\
    w/o SL--CL--CG & $4.43 \pm 0.99$ & 0.14 & 45.92 \\
    
    \bottomrule
\end{tabular}

\label{tab:ablation_attention}

    \end{minipage}
    \hfill
    \begin{minipage}[t]{0.45\linewidth}
        \vspace{0pt}
        \centering
        \includegraphics[width=0.92\linewidth]{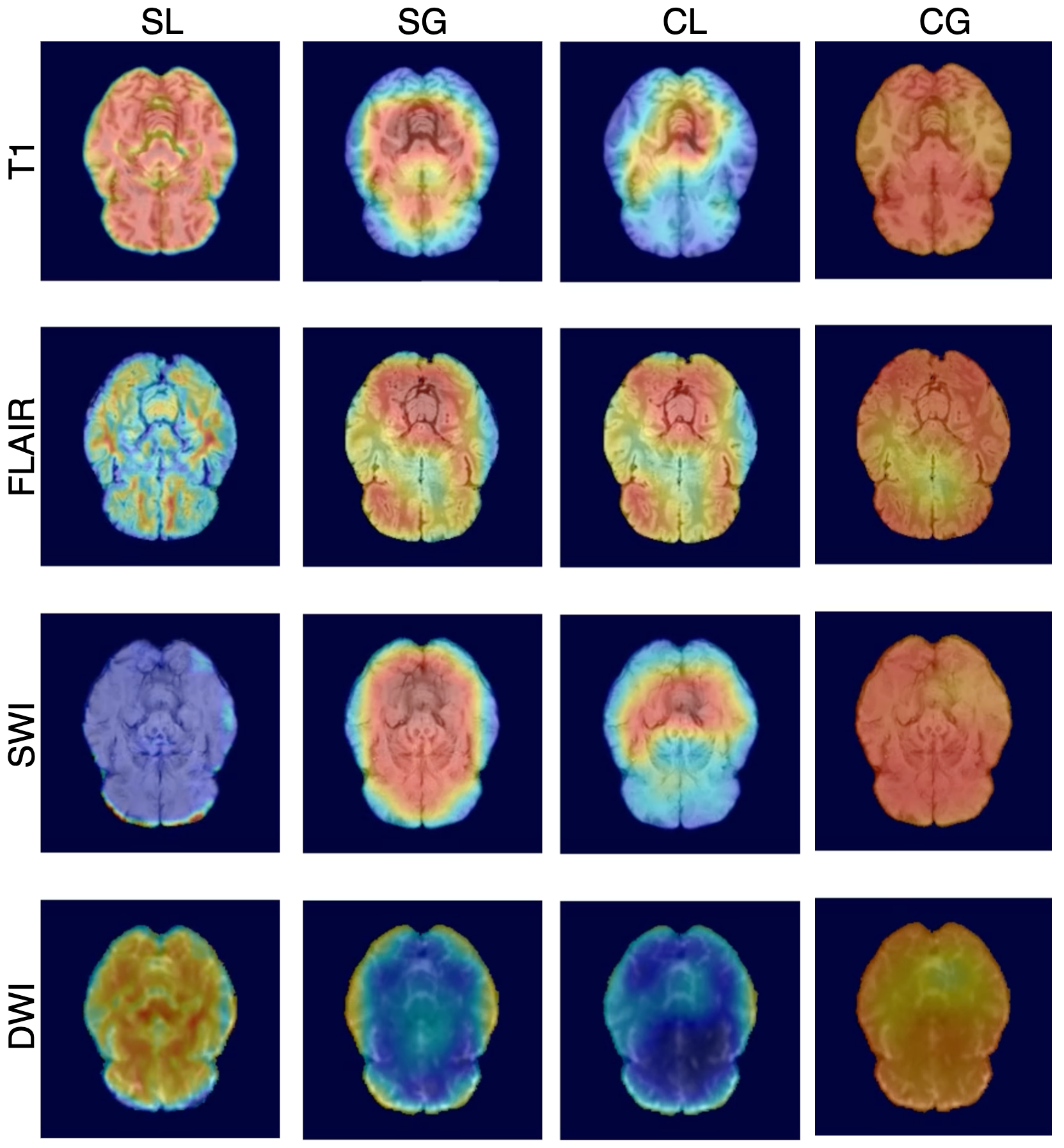}
        \captionof{figure}{
            Grad-CAM \cite{selvarajuGradCAMVisualExplanations2020a} visualizations for \gls{sl}, \gls{sg}, \gls{cl}, and \gls{cg} attention across \gls{t1}, \gls{flair}, \gls{swi}, and \gls{dwi}, averaged over participants in the \gls{ukbb} test set.
            Local and global variants highlight complementary spatial detail and distributed multimodal context.
        }
        \label{fig:attn_maps_2}
    \end{minipage}
\end{figure}

Building on these observations, we perform systematic attention ablations by isolating component contributions using \gls{ukbb} MRI data (\Cref{tab:ablation_attention}). Single attention types consistently underperform, with \gls{sg} yielding the highest error and variance, while \gls{cg} performs substantially better, underscoring the importance of global cross-modal context. Local variants (\gls{sl}, \gls{cl}) provide moderate improvements but remain inferior to \gls{cg} in isolation. Combining attention mechanisms consistently reduces error, with configurations including global attention (\gls{sg}, \gls{cg}) clearly outperforming purely local variants. Performance further improves with higher-order combinations, peaking when all four attention types are used. Adding local attention to global components yields incremental gains, indicating complementary roles in refining spatial detail and enabling cross-modal fusion. Overall, performance scales with attention diversity, with progressively smaller gains as the model approaches the full configuration.

\paragraph{Attribution maps reveal complementary attention across modalities}
The attribution patterns suggest consistent yet complementary roles of the different attention mechanisms across modalities, including patterns associated with brain aging (\Cref{fig:attn_maps_2}).
\gls{sl} captures fine-grained, symmetric anatomical structure and differentiates between gray and white matter. In \gls{t1}, \gls{sl} emphasizes widespread white matter regions, while in \gls{flair} it focuses on subcortical structures with extensions into gyral patterns, as well as brainstem and cerebellar regions, indicating sensitivity to fiber-rich areas that are known to change with age \cite{wardlawNeuroimagingStandardsResearch2013, decarliMeasuresBrainMorphology2005, razRegionalBrainChanges2005, coxAgeingBrainWhite2016}. In contrast, \gls{swi} contributes only marginally at the local level, whereas \gls{dwi} shows increased relevance in white matter and brainstem structures, suggesting complementary microstructural information \cite{beckWhiteMatterMicrostructure2021, farokhianAgeRelatedGrayWhite2017}.
\gls{sg} shifts the focus toward coarse anatomical organization and highlights periventricular regions, posterior brainstem, and cerebellum, structures commonly associated with global atrophy and ventricular enlargement \cite{habesWhiteMatterHyperintensities2016, bouhraraMaturationDegenerationHuman2021, wangMorphologicalChangesCerebellum2024}. \gls{cl} further refines these patterns by aligning spatially consistent regions across modalities. Finally, \gls{cg} produces largely uniform activation across the brain, indicating aggregation of distributed multimodal information, consistent with the view that brain aging manifests as a distributed process across multiple tissue types and spatial scales \cite{chenMultimodalQuantitativeMRI2026, chenMappingHeterogeneousRegion2026}.

\section{Conclusion}

Modeling spatial and cross-modal interactions in 3D brain MRI remains a central challenge. In this work, we introduced \gls{mmgcvit}, a novel architecture that explicitly models intra- and cross-modality dependencies via \gls{sl}, \gls{sg}, \gls{cl}, and \gls{cg} attention, enabling structured interaction learning in both local and global contexts. We showed that \gls{mmgcvit} consistently outperforms established 3D baselines, demonstrating that explicit interaction modeling is a key driver of performance. Moreover, the model benefits more strongly from increasing the number of input modalities than competing architectures, indicating a more effective utilization of complementary information. Across multiple neuroimaging datasets and model scales, \gls{mmgcvit} exhibits robust and consistent performance, indicating stable behavior across varying data conditions and acquisition settings. Despite this strong empirical performance, several limitations remain. First, evaluation is restricted to brain MRI and brain age prediction. Generalization to tasks such as disease classification, segmentation, or prognosis is not yet validated. Second, \gls{mmgcvit} assumes well-aligned multimodal inputs and does not explicitly handle missing, corrupted, or imbalanced modalities. Third, the use of multiple attention mechanisms (\gls{sl}, \gls{sg}, \gls{cl}, \gls{cg}) increases architectural complexity and computational cost, particularly for high-resolution 3D data. Finally, experiments are conducted in a dataset-specific setting. Robustness under stronger domain shifts without explicit cross-dataset adaptation remains an open question. Overall, our results emphasize that explicitly modeling the structure of intra- and cross-modal interactions is critical for effective multimodal learning in 3D medical imaging.

\newpage
\noindent\textbf{Acknowledgment} The authors acknowledge the Scientific Computing of the IT Division at the Charité – Universitätsmedizin Berlin for providing computational resources that contributed to the research results reported in this paper. This research has been conducted using the UK Biobank Resource under Application Number 49966. Parts of the data collection and sharing for this project were provided by the Cambridge Centre for Ageing and Neuroscience (Cam-CAN). Cam-CAN funding was provided by the UK Biotechnology and Biological Sciences Research Council (grant no. BB/H008217/1), together with support from the UK Medical Research Council and the University of Cambridge. We would like to acknowledge the participants, students, faculty, and staff who supported the Stroke Outcome Optimization Project (SOOP) in collecting and curating this dataset.
\bibliographystyle{unsrtnat}
\bibliography{references}
\newpage
\section{Appendix}

\FloatBarrier

\begin{table}[htbp]
    \centering
    \captionsetup{skip=6pt}
    \caption{
Ablation over single-layer attention permutations on \gls{ukbb}. We evaluate all combinations of separated local (\gls{sl}), separated global (\gls{sg}), cross local (\gls{cl}), and cross global (\gls{cg}) attention. The full configuration (\gls{sl}-\gls{sg}-\gls{cl}-\gls{cg}) achieves the best performance (MAE $\downarrow$), consistently outperforming all other permutations. This indicates that gains arise from the interaction of complementary mechanisms rather than any single component. Separated attention (\gls{sl}, \gls{sg}) strengthens intra-modality representations, while cross attention (\gls{cl}, \gls{cg}) enables inter-modality exchange. Orthogonally, local attention (\gls{sl}, \gls{cl}) captures fine-grained structure, whereas global attention (\gls{sg}, \gls{cg}) models long-range dependencies.
}
    \label{tab:single_layer_attention_permutations}
    \begin{tabular}{@{} l c c @{}}
        \toprule
        \textbf{Attention} & \makecell{\textbf{MAE} $\downarrow$\\\textbf{$\pm$ SD}} & \textbf{Parameters} \\
        \midrule
        CL-CL-CL-CL & $4.12 \pm 0.04$ & 176837 \\
        SL-SL-SL-SL & $4.05 \pm 0.06$ & 176837 \\
        CL-SG-CL-SG & $3.80 \pm 0.62$ & 174739 \\
        CL-CG-CL-CG & $3.67 \pm 0.61$ & 174725 \\
        SL-SG-SL-SG & $3.48 \pm 0.22$ & 174725 \\
        CL-CG-CL-SG & $3.47 \pm 0.34$ & 174746 \\
        SL-CG-SL-SG & $3.47 \pm 0.20$ & 174732 \\
        SL-SG-CL-SG & $3.44 \pm 0.10$ & 174725 \\
        SL-SG-SL-CG & $3.41 \pm 0.27$ & 174732 \\
        SG-CL-SG-SL & $3.39 \pm 0.15$ & 174732 \\
        SL-CG-SL-CG & $3.38 \pm 0.16$ & 174739 \\
        SG-SL-SG-SL & $3.38 \pm 0.13$ & 174725 \\
        CG-CL-CG-SL & $3.37 \pm 0.15$ & 174746 \\
        CL-SG-SL-SG & $3.36 \pm 0.36$ & 174725 \\
        CL-SG-SL-CG & $3.36 \pm 0.26$ & 174725 \\
        SL-CG-CL-CG & $3.36 \pm 0.17$ & 174725 \\
        CL-CG-SL-SG & $3.36 \pm 0.36$ & 174725 \\
        SG-SG-SG-SG & $3.36 \pm 0.13$ & 172613 \\
        CL-SG-CL-CG & $3.35 \pm 0.13$ & 174746 \\
        CL-CG-SL-CG & $3.34 \pm 0.26$ & 174725 \\
        SL-CG-CL-SG & $3.33 \pm 0.18$ & 174725 \\
        CG-CL-CG-CL & $3.31 \pm 0.14$ & 174753 \\
        CG-SL-SG-CL & $3.31 \pm 0.17$ & 174725 \\
        CG-CL-SG-SL & $3.31 \pm 0.15$ & 174739 \\
        SG-SL-SG-CL & $3.29 \pm 0.20$ & 174732 \\
        SG-CL-SG-CL & $3.29 \pm 0.10$ & 174739 \\
        CG-CL-SG-CL & $3.26 \pm 0.12$ & 174725 \\
        SG-CL-CG-SL & $3.26 \pm 0.09$ & 174739 \\
        SG-SL-CG-SL & $3.25 \pm 0.04$ & 174732 \\
        SG-SL-CG-CL & $3.24 \pm 0.05$ & 174725 \\
        CG-CG-CG-CG & $3.24 \pm 0.07$ & 172613 \\
        CG-SL-CG-CL & $3.22 \pm 0.04$ & 174746 \\
        CG-SL-SG-SL & $3.21 \pm 0.10$ & 174725 \\
        CG-SL-CG-SL & $3.21 \pm 0.08$ & 174739 \\
        SG-CL-CG-CL & $3.20 \pm 0.09$ & 174746 \\
        \textbf{SL-SG-CL-CG} & $\mathbf{3.12} \pm 0.19$ & 174725 \\
        \bottomrule
    \end{tabular}
\end{table}

\begin{table*}[t]
\centering
\tiny
\setlength{\tabcolsep}{1.5pt}
\renewcommand{\arraystretch}{1.08}
\captionsetup{skip=6pt}
\caption{
Model configurations across modality settings.
Parameters are reported in millions (M).
}
\label{tab:model_configs_modalities_final}

\begin{tabular}{@{}p{0.12\textwidth}p{0.13\textwidth}p{0.17\textwidth}p{0.17\textwidth}p{0.17\textwidth}p{0.17\textwidth}@{}}
\toprule
\textbf{Model} & \textbf{Parameter} 
& \makecell{\textbf{T1/FLAIR/SWI/DWI} \\ modalities=1}
& \makecell{\textbf{FLAIR+DWI} \\ modalities=2} 
& \makecell{\textbf{T1+FLAIR+DWI} \\ modalities=3} 
& \makecell{\textbf{T1+FLAIR+DWI+SWI} \\ modalities=4} \\
\midrule

\multicolumn{6}{l}{\textbf{\gls{mmgcvit} (ours)}} \\
& dim & 224 & 224 & 224 & 224 \\
& num\_heads & 14 & 14 & 14 & 14 \\
& attn\_types 
& [SL, SG, SL, SG] 
& [SL, SG, CL, CG] 
& [SL, SG, CL, CG] 
& [SL, SG, CL, CG] \\
& window\_size & [4,4,4] & [4,4,4] & [4,4,4] & [4,4,4] \\
& Params (M) & 2.28 & 3.55 & 5.04 & 6.97 \\

\midrule

\multicolumn{6}{l}{\textbf{ConvNeXt}} \\
& depths 
& [2,3,4,2] 
& [2,4,10,2] 
& [4,8,10,4] 
& [8,10,12,6] \\
& dims 
& [128,128,128,128] 
& [128,128,128,128] 
& [128,128,128,128] 
& [128,128,128,128] \\
& Params (M) & 2.34 & 3.58 & 5.00 & 6.77 \\

\midrule

\multicolumn{6}{l}{\textbf{GCViT}} \\
& dim & 32 & 32 & 32 & 32 \\
& num\_heads 
& [2,8,8,4] 
& [2,8,8,4] 
& [2,8,8,4] 
& [2,8,8,4] \\
& depth 
& [1,1,1,1] 
& [2,4,4,2] 
& [2,4,5,4] 
& [4,10,14,4] \\
& window\_size 
    & \makecell[l]{[[4,4,4],[4,4,4],\\{}[8,8,8],[4,4,4]]}
    & \makecell[l]{[[4,4,4],[4,4,4],\\{}[8,8,8],[4,4,4]]}
    & \makecell[l]{[[4,4,4],[4,4,4],\\{}[8,8,8],[4,4,4]]}
    & \makecell[l]{[[4,4,4],[4,4,4],\\{}[8,8,8],[4,4,4]]} \\
& Params (M) & 2.28 & 3.56 & 5.01 & 6.96 \\

\bottomrule
\end{tabular}
\end{table*}

\begin{table*}[t]
\centering
\tiny
\setlength{\tabcolsep}{1.5pt}
\renewcommand{\arraystretch}{1.08}
\captionsetup{skip=6pt}
\caption{Model configurations across different scales.}
\label{tab:model_configs}

\begin{tabular}{@{}p{0.12\textwidth}p{0.13\textwidth}p{0.16\textwidth}p{0.16\textwidth}p{0.16\textwidth}p{0.16\textwidth}@{}}
\toprule
\textbf{Model} & \textbf{Parameter} & \textbf{Small} & \textbf{Medium} & \textbf{Base} & \textbf{Large} \\
\midrule

\multicolumn{6}{l}{\textbf{\gls{mmgcvit} (ours)}} \\
& dim & 32 & 64 & 128 & 224 \\
& num\_heads & 2 & 4 & 8 & 14 \\
& attn\_types & [SL, SG, CL, CG] & [SL, SG, CL, CG] & [SL, SG, CL, CG] & [SL, SG, CL, CG] \\
& window\_size & [4,4,4] & [4,4,4] & [4,4,4] & [4,4,4] \\
& Params (M) & 0.17 & 0.62 & 2.34 & 6.97 \\

\midrule
\multicolumn{6}{l}{\textbf{ConvNeXt}} \\
& depths & [1,2,3,2] & [2,2,3,2] & [2,2,4,3] & [8,10,12,6] \\
& dims & [32,32,32,32] & [64,64,64,64] & [128,128,128,128] & [128,128,128,128] \\
& Params (M) & 0.19 & 0.61 & 2.40 & 6.77 \\

\midrule
\multicolumn{6}{l}{\textbf{Swin Transformer}} \\
& embed\_dim & 32 & 64 & 128 & 224 \\
& depths & [3,3] & [2,3] & [2,3] & [1,3] \\
& num\_heads & [2,2] & [4,4] & [4,4] & [4,4] \\
& window\_size & [4,4,4] & [4,4,4] & [4,4,4] & [4,4,4] \\
& Params (M) & 0.17 & 0.62 & 2.45 & 6.79 \\

\midrule
\multicolumn{6}{l}{\textbf{ViT}} \\
& depth & 8 & 12 & 11 & 34 \\
& dim & 32 & 64 & 128 & 128 \\
& heads & 4 & 4 & 8 & 8 \\
& mlp\_dim & 256 & 256 & 512 & 512 \\
& Params (M) & 0.18 & 0.74 & 2.71 & 6.87 \\

\midrule
\multicolumn{6}{l}{\textbf{GCViT}} \\
& depth & [1,2,2,1] & [2,2,6,2] & [2,6,6,2] & [4,10,14,4] \\
& num\_heads & [2,2,2,2] & [2,2,4,2] & [2,4,6,2] & [2,8,8,4] \\
& dim & 8 & 12 & 24 & 32 \\
    & window\_size
    & \makecell[l]{[[4,4,4],[4,4,4],\\{}[8,8,8],[4,4,4]]}
    & \makecell[l]{[[4,4,4],[4,4,4],\\{}[8,8,8],[4,4,4]]}
    & \makecell[l]{[[4,4,4],[4,4,4],\\{}[8,8,8],[4,4,4]]}
    & \makecell[l]{[[4,4,4],[4,4,4],\\{}[8,8,8],[4,4,4]]} \\
& Params (M) & 0.16 & 0.62 & 2.31 & 6.96 \\

\midrule
\multicolumn{6}{l}{\textbf{ResNet}} \\
& dim & [8,16,32,64] & [16,32,64,64] & [16,32,64,128] & [16,32,64,128] \\
& layers & [1,1,1,1] & [2,2,2,2] & [2,4,6,3] & [12,20,32,4] \\
& strides & [1,2,2,1] & [1,2,2,1] & [1,2,2,1] & [1,2,2,1] \\
& Params (M) & 0.18 & 0.60 & 2.46 & 6.88 \\

\midrule
\multicolumn{6}{l}{\textbf{ViT (Late Fusion)}} \\
& depth & 2 & 2 & 2 & 4 \\
& dim & 32 & 64 & 128 & 176 \\
& heads & 2 & 4 & 8 & 8 \\
& dim\_fusion & 32 & 64 & 128 & 176 \\
& heads\_fusion & 2 & 4 & 8 & 16 \\
& Params (M) & 0.19 & 0.54 & 2.12 & 6.98 \\

\midrule
\multicolumn{6}{l}{\textbf{ResNet (Late Fusion)}} \\
& hidden\_dims & [8,12,14,16] & [8,16,20,32] & [8,16,32,64] & [16,32,64,88] \\
& resnet\_layers & [2,2,2,2] & [2,3,4,2] & [2,4,8,2] & [4,5,8,2] \\
& strides & [1,2,2,1] & [1,2,2,1] & [1,2,2,1] & [1,2,2,1] \\
& dim\_fusion & 32 & 64 & 128 & 176 \\
& heads\_fusion & 2 & 4 & 8 & 16 \\
& Params (M) & 0.18 & 0.60 & 2.28 & 7.00 \\

\bottomrule
\end{tabular}
\end{table*}

\begin{table}[t]
\centering
\setlength{\tabcolsep}{4pt}
\captionsetup{skip=6pt}
\caption{
Scaling comparison on UKBB. Mean Absolute Error (MAE) in years ($\pm$SD).
Late fusion variants are marked with $\ddag$. Best results in \textbf{bold}, second best \underline{underlined}.
}
\label{tab:scaling_results}
\begin{tabular}{l l r r r}
\toprule
\textbf{Model} & \textbf{Config} & \textbf{Params (M)} & \makecell{\textbf{Traing Time}\\\textbf{(min/epoch)}} & \makecell{\textbf{MAE} $\downarrow$\\\textbf{$\pm$ SD}} \\
\midrule

ConvNeXt
& Small  & 0.19 & 21.57  & $3.526 \pm 0.082$ \\
& Medium & 0.61 & 60.48  & $3.484 \pm 0.037$ \\
& Base   & 2.40 & 67.07  & $3.464 \pm 0.040$ \\
& Large  & 6.77 & 67.67 & $3.447 \pm 0.039$ \\

\midrule

ResNet
& Small  & 0.18 & 26.64  & $4.613 \pm 0.687$ \\
& Medium & 0.60 & 38.35  & $4.060 \pm 0.068$ \\
& Base   & 2.46 & 38.43  & $3.935 \pm 0.068$ \\
& Large  & 6.88 & 34.50  & $3.711 \pm 0.168$ \\

ResNet $\ddag$
& Small  & 0.18 & 22.52  & $5.176 \pm 2.364$ \\
& Medium & 0.60 & 28.05  & $3.605 \pm 0.377$ \\
& Base   & 2.28 & 34.26  & $3.570 \pm 0.105$ \\
& Large  & 7.00 & 42.07  & $3.443 \pm 0.123$ \\

\midrule

Swin-T
& Small  & 0.17 & 55.18   & $4.005 \pm 0.203$ \\
& Medium & 0.62 & 57.41   & $3.742 \pm 0.159$ \\
& Base   & 2.45 & 45.12   & $3.665 \pm 0.137$ \\
& Large  & 6.79 & 64.76   & $3.536 \pm 0.076$ \\

\midrule

GCViT
& Small  & 0.16 & 43.81  & $4.094 \pm 0.617$ \\
& Medium & 0.62 & 58.54  & $3.453 \pm 0.101$ \\
& Base   & 2.31 & 68.20  & $3.410 \pm 0.034$ \\
& Large  & 6.96 & 89.45  & $3.342 \pm 0.065$ \\

\midrule

ViT
& Small  & 0.18 & 48.92  & $4.511 \pm 0.781$ \\
& Medium & 0.74 & 90.11  & $4.097 \pm 0.183$ \\
& Base   & 2.71 & 91.53  & $4.006 \pm 0.205$ \\
& Large  & 6.87 & 96.82 & $3.832 \pm 0.146$ \\

ViT $\ddag$
& Small  & 0.19 & 64.56  & $4.878 \pm 1.158$ \\
& Medium & 0.54 & 52.05  & $4.376 \pm 0.487$ \\
& Base   & 2.12 & 53.02  & $3.976 \pm 0.227$ \\
& Large  & 6.98 & 84.56  & $3.677 \pm 0.066$ \\

\midrule

\textbf{\gls{mmgcvit}}
& Small  & 0.17 & 38.84   & $3.119 \pm 0.191$ \\
& Medium & 0.62 & 60.37   & $3.000 \pm 0.076$ \\
& Base   & 2.34 & 69.54   & $\underline{2.892} \pm 0.035$ \\
& Large  & 6.97 & 86.33   & $\mathbf{2.816} \pm 0.062$ \\

\bottomrule
\end{tabular}
\end{table}

\begin{table}[t]
\centering
\scriptsize
\setlength{\tabcolsep}{3pt}
\renewcommand{\arraystretch}{1.25}
\captionsetup{skip=6pt}
\caption{Hyperparameter search space for ConvNeXt across model scales.}
\label{tab:hyperparams_convnext}
\begin{tabular}{@{}l l l l l l@{}}
\toprule
\textbf{Section} & \textbf{Parameter} & \textbf{Small} & \textbf{Medium} & \textbf{Base} & \textbf{Large} \\
\midrule



\multicolumn{6}{@{}l}{\textbf{ConvNeXt}} \\
& \texttt{depths} 
& \texttt{[1,2,3,2]} 
& \texttt{[2,2,3,2]} 
& \texttt{[2,2,14,2]} 
& \texttt{[6,8,10,12]} \\

& \texttt{dims} 
& \texttt{[32,32,32,32]} 
& \texttt{[64,64,64,64]} 
& \texttt{[96,96,96,96]} 
& \texttt{[128,128,128,128]} \\

& \texttt{droppath} 
& \multicolumn{4}{l}{\texttt{[0.0, 0.1, 0.2, 0.3]}} \\

\addlinespace[4pt]
\midrule

\multicolumn{6}{@{}l}{\textbf{Optimizer}} \\
& \texttt{learning\_rate} 
& \multicolumn{4}{l}{
\makecell[l]{min: \texttt{1e\text{-}5}, max: \texttt{5e\text{-}3}\\log-uniform}
} \\

& \texttt{weight\_decay} 
& \multicolumn{4}{l}{
\makecell[l]{min: \texttt{1e\text{-}3}, max: \texttt{3e\text{-}1}\\log-uniform}
} \\

\bottomrule
\end{tabular}
\end{table}

\FloatBarrier

\subsection{MRI preprocessing}
\label{sec:mri_preprocessing}
The MRI data were preprocessed using a standardized pipeline implemented in Python 3.11 with Advanced Normalization Tools (ANTs). First, raw Digital Imaging and Communications in Medicine (DICOM) images were converted to Neuroimaging Informatics Technology Initiative (NIfTI) format to ensure compatibility with downstream processing. Noise reduction was applied to improve the signal-to-noise ratio, followed by bias field correction to compensate for intensity inhomogeneities. Skull stripping was then performed to remove non-brain tissue and isolate intracranial structures. Subsequently, intensity normalization was applied to harmonize voxel intensity distributions across subjects. Finally, all images were spatially aligned to a common anatomical reference space via co-registration to the Montreal Neurological Institute (MNI) space, enabling voxel-wise comparability across the cohort. Given the high computational complexity of volumetric MRI data, all scans were initially downsampled to a resolution of $64 \times 64 \times 64$ voxels from their original resolution of $224^3$.

\subsection{Attention computation}
Given an input sequence of 3D patch embeddings $Z \in \mathbb{R}^{N \times C}$, where $N$ represents the number of tokens and $C$ is the embedding dimension, embeddings are projected into query ($Q$), key ($K$), and value ($V$) matrices. This is done using learned linear transformations $W_Q^h, W_K^h, W_V^h \in \mathbb{R}^{C \times d_h}$ for each of the $H$ attention heads. For a single head $h$, the matrices are:

\begin{equation}
Q_h = Z W_Q^h, \quad K_h = Z W_K^h, \quad V_h = Z W_V^h
\end{equation}

where $Q_h, K_h, V_h \in \mathbb{R}^{N \times d_h}$ and $d_h = C/H$ is the dimension per head.

The attention scores for each head are computed using scaled dot-product attention. However, to make the model aware of the 3D spatial relationships, we integrate relative positional encodings (RPE) by adding a learnable bias term $B_h \in \mathbb{R}^{N \times N}$ to the attention scores before the softmax normalization. The attention output for a single head is thus calculated as:

\begin{equation}
\text{Attention}(Q_h, K_h, V_h) = \text{softmax} \left( \frac{Q_h K_h^\top}{\sqrt{d_h}} + B_h \right) V_h
\end{equation}

The outputs from all $H$ attention heads are concatenated and passed through a final linear projection layer $(W_O \in \mathbb{R}^{C \times C})$ to produce the final output of the MHSA module:

\begin{equation}
\text{MHSA}(Z) = \text{Concat}(\text{head}_1, \dots, \text{head}_H) W_O
\end{equation}

This mechanism allows a 3D attention-based architecture to effectively model both local and global spatial interactions, informed by the relative 3D positions of the tokens.

\subsection{Implementation Details of MICViT}
Let $x \in \mathbb{R}^{B \times M \times H \times W \times D}$ denote a batch of volumetric inputs with $B$ samples, $M$ MRI modalities, and spatial resolution $H \times W \times D$. Our \gls{mmgcvit} extends the vanilla Global Context Vision Transformer to jointly model modality-specific structure and cross-modal interactions in 3D neuroimaging.

Specifically, each modality is first processed independently by a modality-specific 3D patch embedding operator with kernel size $3$ and stride $2$, producing an initial token tensor
\[
z^{(0)} \in \mathbb{R}^{B \times M \times H_0 \times W_0 \times D_0 \times C_0},
\]
where $C_0 = \mathrm{dim}$ and $(H_0, W_0, D_0)$ denote the patch-level spatial resolution. The token sequence is then passed through a four-stage hierarchical encoder. Let $l \in \{1,2,3,4\}$ denote the stage index; the representation after stage $l$ is
\[
z^{(l)} \in \mathbb{R}^{B \times M \times H_l \times W_l \times D_l \times C_l},
\]
with channel dimensionality increasing as $C_l = 2^{\,l-1} C_0$, while spatial dimensions are progressively reduced by modality-wise downsampling blocks. Each downsampling block combines depthwise 3D convolution, squeeze-and-excitation style channel recalibration, pointwise projection, and strided volumetric reduction, thereby preserving local inductive bias while building a multi-scale token hierarchy.

Within each stage, \gls{mmgcvit} alternates between four attention mechanisms designed to disentangle intra-modality and inter-modality dependencies at both local and global scales: separated local (sl), separated global (sg), cross local (cl), and cross global (cg) attention. Local attention is performed within non-overlapping 3D windows of size $[4,4,4]$, enabling efficient modeling of fine-grained anatomical patterns, while global attention injects long-range context via learned global query tokens.

In contrast to vanilla GCViT, which derives a single global query per stage, our multimodal design employs two complementary query generators: (i) a separated global query generator that extracts modality-specific global context independently per modality, and (ii) a cross-modal global query generator that fuses modalities along the channel dimension and projects the resulting shared representation back into the per-modality attention space. Formally, at stage $l$, these modules produce $q_{\mathrm{sep}}^{(l)}$ and $q_{\mathrm{cross}}^{(l)}$, which parameterize the sg and cg blocks, respectively. This design enables the network to retain modality-specialized representations while explicitly modeling cross-modal dependencies and complementary information.

Each transformer block follows a standard residual structure consisting of normalization, window-based attention, residual connection, and MLP update, with stochastic depth applied across layers. After the final stage, the representation
\[
z^{(4)} \in \mathbb{R}^{B \times M \times H_4 \times W_4 \times D_4 \times C_4}
\]
is normalized and reshaped by merging the modality and channel dimensions to obtain
\[
\tilde{z} \in \mathbb{R}^{B \times (M C_4) \times H_4 \times W_4 \times D_4}.
\]
Global average pooling yields a compact subject-level embedding $h \in \mathbb{R}^{B \times M C_4}$, which is fed into a multitask prediction head.

Across model scales, we vary the base embedding width and the number of attention heads while keeping the architectural template fixed: \gls{mmgcvit}-Small uses $(\mathrm{dim}, \mathrm{num\_heads}) = (32, 2)$, \gls{mmgcvit}-Medium uses $(64, 4)$, \gls{mmgcvit}-Base uses $(128, 8)$, and \gls{mmgcvit}-Large uses $(224, 14)$. All variants use $\texttt{attn\_types} = [\mathrm{sl}, \mathrm{sg}, \mathrm{cl}, \mathrm{cg}]$, $\texttt{patch\_size} = 3$, $\texttt{patch\_stride} = 2$, and $\texttt{window\_size} = [4,4,4]$.

\subsection{Computational Complexity}
The computational complexity of \gls{mmgcvit} is governed by its hierarchical 3D tokenization and window-based attention. At stage $i$, let $M$ denote the number of modalities, $N_i$ the number of tokens per modality, $C_i$ the feature dimension, and $W_i = w_i^{(d)} w_i^{(h)} w_i^{(w)}$ the 3D window volume. Projection and feed-forward layers scale as $\mathcal{O}(M N_i C_i^2)$.

Modality-separated attention operates on intra-modality windows with complexity $\mathcal{O}(M N_i W_i C_i)$, which is linear in $N_i$ for fixed window size. Cross-modal attention jointly attends across modalities within each window and increases the cost to $\mathcal{O}(M^2 N_i W_i C_i)$ through pairwise modality interactions. Global attention uses stage-wise global queries with window-local keys and values, avoiding quadratic scaling in the total number of tokens.

Aggregating over all stages yields
\begin{equation}
\mathcal{O}\left(
\sum_i \left[
(S_i + X_i) M N_i C_i^2
+ S_i M N_i W_i C_i
+ X_i M^2 N_i W_i C_i
\right]
\right),
\end{equation}
where $S_i$ and $X_i$ denote the numbers of modality-separated and cross-modal attention blocks at stage~$i$, respectively. Overall, \gls{mmgcvit} scales linearly in the number of tokens $N_i$ (for fixed window size), in contrast to full volumetric self-attention, which scales quadratically as $\mathcal{O}(M^2 N_i^2 C_i)$. This favorable scaling is critical for enabling high-resolution multimodal 3D neuroimaging.

\subsection{Multi-modal patch embedding.}
\label{sec:stem}
Let $\mathbf{x} \in \mathbb{R}^{B \times M \times H \times W \times D}$ denote a batch of $M$ co-registered MRI modalities. We first map each modality to a latent patch representation using a modality-specific 3D patch embedding operator. Specifically, for modality $m$, we extract $\mathbf{x}^{(m)} \in \mathbb{R}^{B \times 1 \times H \times W \times D}$ and compute
\[
\mathbf{z}^{(m)} = f_m\!\left(\mathbf{x}^{(m)}\right) \in \mathbb{R}^{B \times H' \times W' \times D' \times C},
\]
where $f_m$ consists of a strided 3D convolution with kernel size $P$ and stride $S$, followed by a local convolutional refinement block. The strided convolution performs patchification and linear projection into a $C$-dimensional embedding space, while the refinement block adds an inductive bias for local anatomical continuity through depthwise 3D convolution, nonlinearity, squeeze-and-excitation, and pointwise mixing. Importantly, this refinement block does not further reduce spatial resolution at the embedding stage.

The modality-specific embeddings are then stacked as
\[
\mathbf{z} = \mathrm{stack}\left(\mathbf{z}^{(1)}, \dots, \mathbf{z}^{(M)}\right)
\in \mathbb{R}^{B \times M \times H' \times W' \times D' \times C}.
\]
This design preserves an explicit modality axis, rather than collapsing modalities into channels at the input. As a result, subsequent transformer blocks can model intra-modality structure and inter-modality interactions separately, which is particularly desirable for multi-contrast MRI where complementary information is distributed across modalities. Note that the stem layers are not identical across all models and baselines. While MICViT employs modality-specific patch embeddings with convolutional refinement, baseline architectures use their respective standard input stems, reflecting their underlying design assumptions.

\subsection{Details of resources used}
\label{sec:resources}

Our experiments are conducted on a High-Performance Cluster (HPC) with the following environment. We use 80GB NVIDIA H100 and NVIDIA H200 GPUs for training and evaluation. The cluster nodes are equipped with Intel Xeon Platinum (8468 or 8568Y+) or AMD EPYC (9534 or 9355) CPUs and 754GB–2TB of RAM.

\end{document}